\pdfoutput=1

\documentclass[11pt]{article}

\usepackage[final]{acl}

\usepackage{times}
\usepackage{latexsym}
\usepackage[symbol]{footmisc}
\usepackage{subfigure}
\usepackage{float}
\usepackage{graphicx}
\usepackage{booktabs}
\usepackage{amsmath}

\DeclareMathOperator*{\argmin}{arg\,min}

\usepackage{enumitem}

\usepackage[T1]{fontenc}
\usepackage[utf8]{inputenc}
\usepackage{microtype}
\usepackage{inconsolata}

\usepackage{graphicx}

\title{Scaling Laws Across Model Architectures: A Comparative Analysis of Dense and MoE Models in Large Language Models}

\author{
 \textbf{Siqi Wang\textsuperscript{1,2}},
 \textbf{Zhengyu Chen\textsuperscript{1}\footnotemark[1]},
 \textbf{Bei Li\textsuperscript{1}},
 \textbf{Keqing He\textsuperscript{1}},
\\
 \textbf{Min Zhang\textsuperscript{2}},
 \textbf{Jingang Wang\textsuperscript{1} \footnotemark[1]}
\\
\\
 \textsuperscript{1}Meituan Inc.
 \textsuperscript{2}The University of Hong Kong
\\
\quad\texttt{\{chenzhengyu04,libei17,hekeqing,wangjingang02\}@meituan.com}\\
\quad\texttt{\{siqiwang,minzhang12\}@hku.hk}\\
}
\begin{document}
\maketitle

\footnotetext[1]{Corresponding authors.}

\begin{abstract}
The scaling of large language models (LLMs) is a critical research area for the efficiency and effectiveness of model training and deployment. Our work investigates the transferability and discrepancies of scaling laws between Dense Models and Mixture of Experts (MoE) models. Through a combination of theoretical analysis and extensive experiments, including consistent loss scaling, optimal batch size and learning rate scaling, and resource allocation strategies scaling, our findings reveal that the power-law scaling framework also applies to MoE Models, indicating that the fundamental principles governing the scaling behavior of these models are preserved, even though the architecture differs. 
Additionally, MoE Models demonstrate superior generalization, resulting in lower testing losses with the same training compute budget compared to Dense Models. These findings indicate the scaling consistency and transfer generalization capabilities of MoE Models, providing new insights for optimizing MoE Model training and deployment strategies.

\end{abstract}

\section{Introduction}
The advent and scaling of large language models (LLMs), such as GPT \citep{brown2020language, achiam2023gpt}, Llama \citep{touvron2023llamaa, touvron2023llamab}, Gemini \citep{team2023gemini}, Gopher \citep{rae2021scaling}, Chinchilla \citep{hoffmann2022training}, and Mistral \citep{jiang2023mistral}, have marked a transformative era in artificial intelligence and natural language processing. Characterized by their vast parameter counts and extensive training datasets, these models have significantly advanced capabilities across various domains, including machine translation \citep{brown2020language, Hendy2023HowGA, garcia2022using}, logical reasoning \citep{huang2022towards,wei2022chain,chen2021pareto,chen2024learning}, and medical applications \citep{thirunavukarasu2023large,xiao2022decoupled}. However, their increasing complexity and parameter scale posits an urgent need for innovative scaling strategies that optimize computational efficiency without compromising performance.

Historically, Dense Transformer Models have dominated due to their simplicity and scalability. And the scaling laws \citep{kaplan2020scaling, hoffmann2022training} for dense models have been thoroughly investigated across different circumstances, such as over-training \citep{gadre2024language} and data-limiting \citep{muennighoff2024scaling,chen2022ba}. Despite their efficacy, the huge computational demands of these models necessitate exploration of alternative architectures like Mixture of Experts (MoE) \citep{yuksel2012twenty, shazeer2017outrageously, du2022glam, shen2024jetmoe,chen2021deep,xiao2021learning}, which offer a promising reduction in computational load through sparse activations and dynamic expert routing.

This paper delves into the analysis of scaling laws for Dense and MoE Models within the context of LLMs. We extend foundational research on hyperparameters, such as compute budget, batch size, and learning rate \citep{kaplan2020scaling, hoffmann2022training, mccandlish2018empirical, li2024surge,chen2024pareto}, to explore their transferability and applicability across these architectures. Our experiments involve models up to 7 billion parameters and datasets exceeding 100 billion tokens, aiming to uncover universal scaling behaviors potentially applicable to both model types.

Our results verify the hypothesis that certain scaling laws, particularly those related to loss and hyperparameters, may indeed be universal, bridging architectural gaps between Dense and MoE Models. This universality suggests a simplification in hyperparameter tuning across different scales and architectures, which could significantly streamline the training processes for various LLMs. Furthermore, we provide detailed analyses of the differential impacts of coefficient changes between MoE and Dense Models, offering both empirical and theoretical insights into the superior data efficiency of MoE Models. Concretely, MoE Models can achieve comparable performance with fewer training tokens than Dense Models, alleviating data constraints in LLM training.
Our findings can be summarized as follows:

\begin{itemize}
    \item Consistent Scaling Law Framework: Both MoE and Dense Models demonstrate a consistent and transferable scaling law framework, encompassing loss scaling as well as optimal batch size and learning rate scaling. This alignment implies that the established practices and insights for optimizing Dense Models can be readily applied to MoE Models, potentially streamlining the process of identifying optimal hyperparameters and reducing experimental complexity.
    
    \item Enhanced Data Efficiency in MoE Models: MoE Models demonstrate an approximate 16.37\% improvement in data utilization over Dense Models under similar computational budgets. Theoretical and empirical analyses suggest that during training process, MoE Models, particularly when utilizing the Adam Optimizer, experience lower gradient noise scales. These results show that MoE Models could achieve stable training with smaller batch sizes and larger learning rates, potentially speeding up the training process and improving training convergence. 
    
\end{itemize}

\section{Related Work}

\paragraph{Large Language Models}
Large language models (LLMs) such as GPT \citep{brown2020language}, Llama \citep{touvron2023llamaa, touvron2023llamab}, Chinchilla \citep{hoffmann2022training}, Gopher \citep{rae2021scaling}, Mixtral 8x7B \citep{jiang2024mixtral}, Switch Transformer \citep{fedus2022switch}, GLaM \citep{du2022glam}, and DeepSpeed-MoE \citep{rajbhandari2022deepspeed} have advanced significantly, categorized into Dense Models and Mixture of Experts (MoE) Models. Dense Models activate all parameters per forward pass, while MoE Models activate only a subset, allowing for larger model scales without proportional increases in computational costs. Despite their complexity, MoE Models have shown potential for superior performance and efficiency.
\paragraph{Scaling Laws for LLMs}
Due to the significant costs associated with training process, understanding the scaling laws of large language models (LLMs) is crucial. Studies \citep{bahri2021explaining, kaplan2020scaling,bi2024deepseek} have established a power-law relationship between model loss and factors like training tokens and compute budget. Recent work \citet{Yun2024TowardIM} has explored these relationships further in MoE Models, indicating cost-effective scaling benefits but also highlighting challenges such as expert selection and load balancing.
However, a systematic investigation into the scaling laws of MoE Models' hyperparameters and the transferability of scaling laws between Dense Models and MoE Models remains lacking, which is the focus of our work.

\paragraph{Hyperparameters Estimation}
As model sizes increase, precise optimal hyperparameter estimation becomes critical \cite{chen2021multi}. Research \citet{mccandlish2018empirical} has focused on optimizing batch size and learning rates to balance training speed and efficiency. Novel approaches \citep{yang2022tensor, yang2023tensor} like Maximal Update Parametrization suggest that optimal hyperparameters for smaller models might scale to larger models effectively. Our study extends these insights to explore hyperparameter transferability between Dense and MoE Models, focusing on resource allocation, learning rate, batch size, and their transfer rules for Dense Models and MoE Models.

\section{Preliminary}

\begin{figure}[t]
	\centering
	\vspace{-0.1in}        \includegraphics[width=0.98\linewidth,height=2.1in]{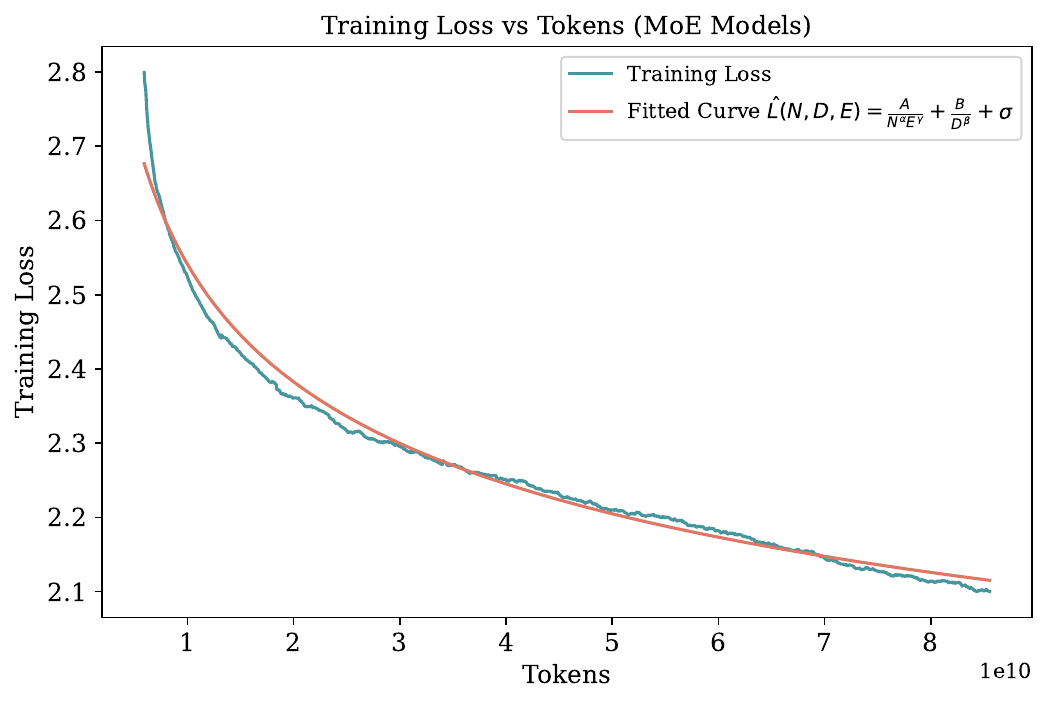}	
	\vspace{-0.1in}	
	\caption{The extrapolated scaling curves for 1.5B Mixture of Experts (MoE) models. This demonstrates that the proposed Loss Scaling Curve \( \hat{L}(N,D,E) = \frac{A}{N^\alpha E^\gamma} + \frac{B}{D^\beta} +\sigma (E \textless 100 )\), fits well for MoE Models (eight experts). Specifically, \( D \) is the number of tokens and \( N \) is the model scale, which is compute budget (\( C \)) divided by \( D \), instead of the model size. \( E \) is the number of experts and \( \sigma \) represents the random noise scale of dataset. $A$, $B$, $\gamma$, $\alpha$ and $\beta$ are all coefficients.}
	\vspace{-0.1in}
	\label{fig:training_loss_extrapolation_1500m}
\end{figure}

The scaling law of the training loss for Dense Models with respect to the number of training tokens and model size has been extensively studied \citep{kaplan2020scaling, hoffmann2022training}. Previous work \citep{hoffmann2022training} has proposed the following scaling law (shown in Equation~\ref{eq:L_D_N_sigma}). To introduce the concept of model scale (the FLOPs divided by the number of training tokens) as $N$ in our work, we denote model size (number of parameters) as $P$ to avoid confusion.
\begin{equation}
\hat{L}(P, D) = \frac{A}{P^\alpha} + \frac{B}{D^\beta} + \sigma
\label{eq:L_D_N_sigma}
\end{equation}
\[
\text{s.t.} \quad \text{FLOPs}(P, D) = C
\]
\noindent where $\hat{L}$ is the training loss, $D$ is the number of training tokens, $P$ is the model size (number of parameters), and $\sigma$ represents the minimum achievable training loss due to the dataset's inherent noise. $C$ is the compute budget with an approximation $C = 6PD$. $\alpha$, $\beta$, $A$ and $B$ are all coefficients.

For MoE (Mixture of Experts) Models, previous research \citep{pmlr-v162-clark22a} has proposed a separable scaling law (Equation~\ref{eq:separated_power_law}) for loss between model size and the number of experts.
\begin{equation}
\hat{L}(P, E) = a \log(P) + b \log(E) + d
\label{eq:separated_power_law}
\end{equation}
\noindent where $P$ is the number of model parameters, $E$ is the number of experts, and $a, b, d$ are coefficients. This equals to Equation~\ref{eq:separated_power_law_non_log}.
\begin{align}
    \hat{L}(P, E) &= \frac{10^d}{P^a E^b}
    \label{eq:separated_power_law_non_log}
\end{align}
When using Equation~\ref{eq:separated_power_law} to fit the training loss curve of MoE Models, \citet{pmlr-v162-clark22a} claim that a decrease in performance has been observed given by expert scaling. Specifically, the value of $b$ increases with model size (in Equation~\ref{eq:separated_power_law}) when model size is large. This suggests that as the model size increases, the benefit from increasing the number of experts $E$ will finally decrease. To enhance the fitting ability of the scaling laws for MoE Models, a quadratic interaction term is added, resulting in Equation~\ref{eq:quadratic_interaction_law}.
\begin{align}
    \hat{L}(P, E) &= a \log(P) + b \log(E)  \nonumber \\
    &\quad + c \log(P) \log(E) + d \nonumber \\ 
    &= \frac{10^d}{P^a E^{b+clog(P)}} =  \frac{10^d}{E^b P^{a+clog(E)}}
    \label{eq:quadratic_interaction_law}
\end{align}

\section{Estimating Resources Allocation Strategy Scaling}

\begin{figure*}[t]
	\centering
	\vspace{-0.1in}
	\begin{minipage}{1\linewidth}
            \subfigure[]{
			\label{fig:heatmap_100m_4.0e-03_2.0e-04}			\includegraphics[width=0.48\linewidth,height=2.1in]{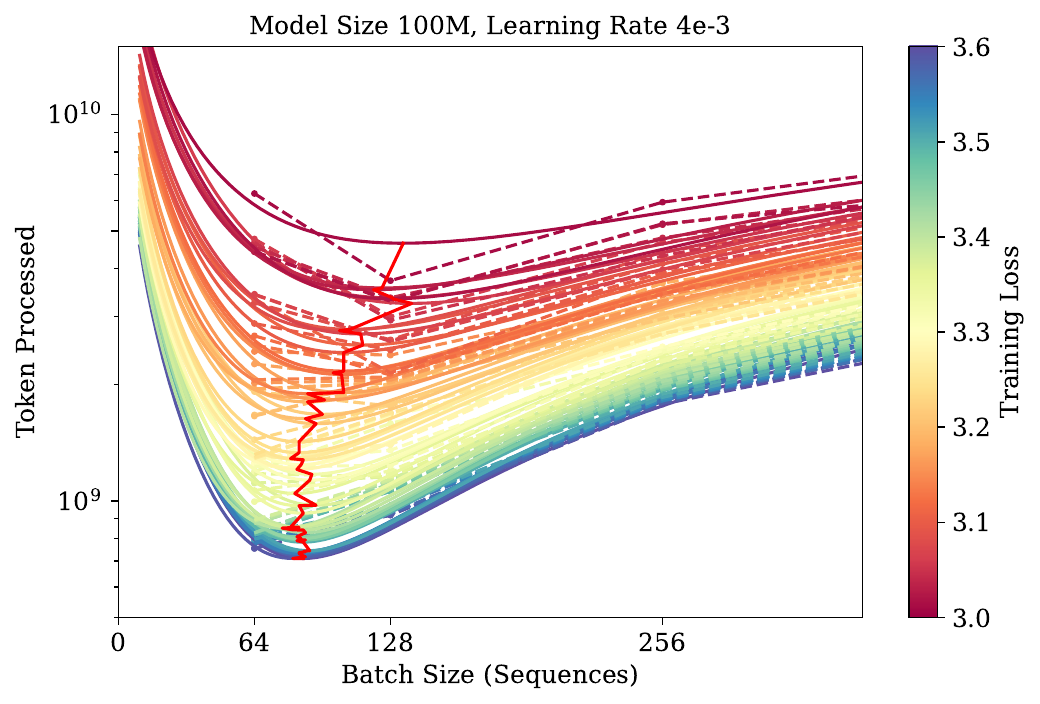}	
		}
            \subfigure[]{
			\label{fig:heatmap_700m_1.0e-03_5.0e-05}			\includegraphics[width=0.48\linewidth,height=2.1in]{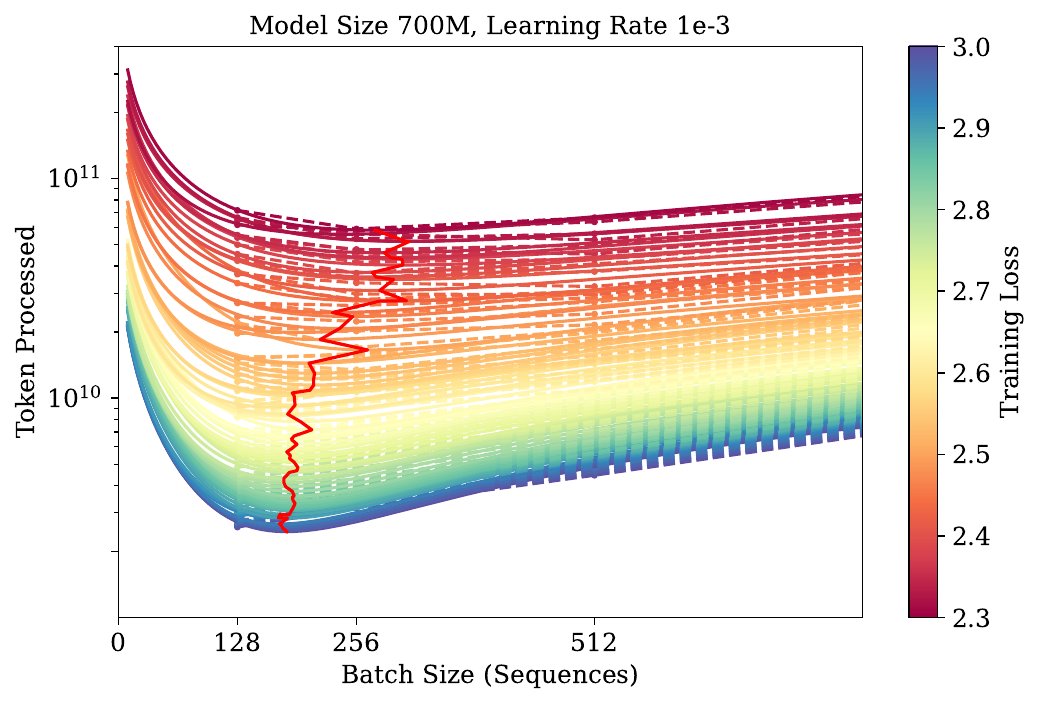}	
		}\noindent
	\end{minipage}
	\vspace{-0.1in}	
	\caption{This diagram presents a heatmap of the distribution of training loss in relation to optimal batch size and training token quantities, with fitted curves representing different training loss. A vertical red line connects the minimum values of each curve. (a) training loss vs. optimal batch size when the MoE model size is 100M and the learning rate is 4e-3. (b) training loss vs. optimal batch size when the MoE model size is 700M and the learning rate is 1e-3.}
 
	\vspace{-0.1in}
	\label{fig:heatmap_bs_loss}
\end{figure*}

\begin{figure*}[ht]
	\centering
	\vspace{-0.1in}
	\begin{minipage}{1\linewidth}
            \subfigure[]{
			\label{fig:optimal_bs_loss_dense}			\includegraphics[width=0.48\linewidth,height=2.2in]{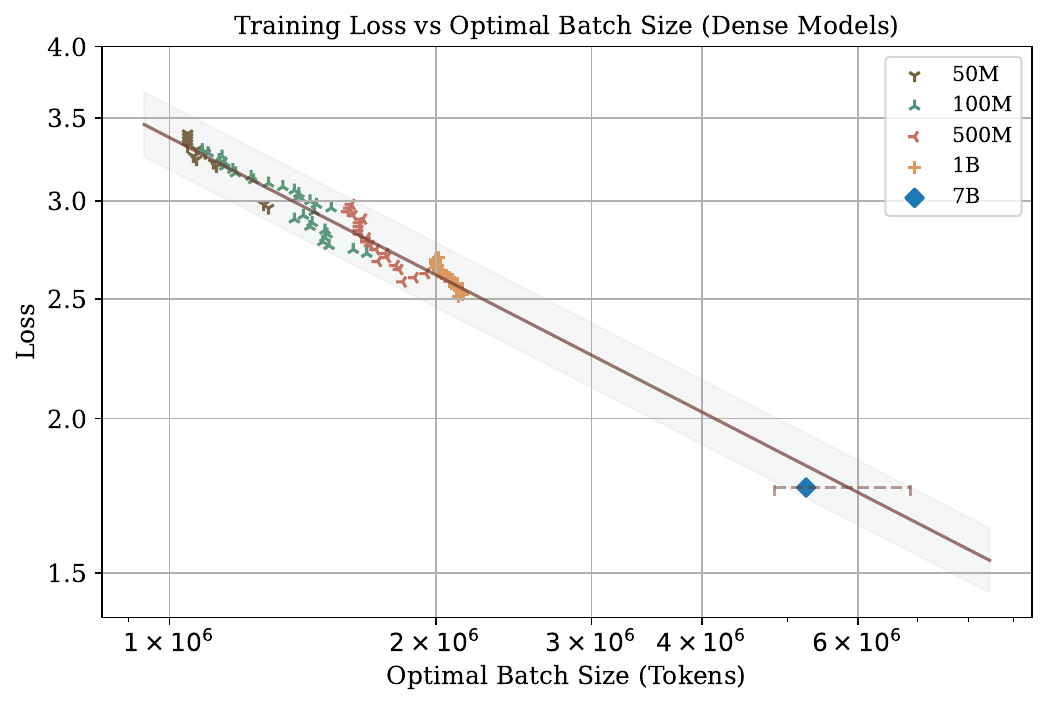}	
		}\noindent
            \subfigure[]{
			\label{fig:optimal_bs_loss_MoE}			\includegraphics[width=0.48\linewidth,height=2.2in]{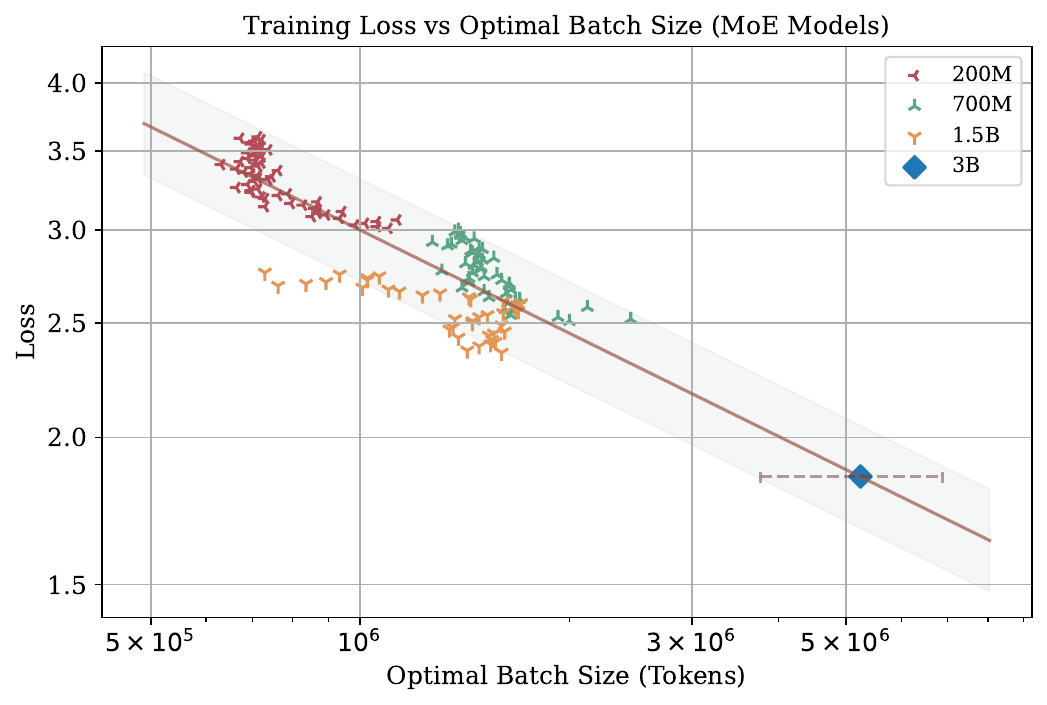}	
		}\noindent
	\end{minipage}
	\vspace{-0.1in}	
	\caption{We plot the optimal batch size values together with the corresponding training loss values across different model sizes for both Dense Models and MoE Models. In log scale diagrams, (a) demonstrates the log-log relationship of training loss vs. optimal batch size for Dense Models. (b) demonstrates the log-log relationship of training loss vs. optimal batch size for MoE Models. This indicates that the power-law relationship remains consistent not only across model sizes but also across model architectures. The total overlap of the comparative performance interval is about 65.8\%.}
	\vspace{-0.in}
	\label{fig:transfer_bs_loss}
\end{figure*}
\subsection{Scaling Laws for Training Loss}
\citet{kaplan2020scaling} and \citet{hoffmann2022training} originally proposed a scaling law for Dense Models, while \citet{pmlr-v162-clark22a} extended this law to scenarios involving multiple experts (MoE Models). Upon closer examination of Equation~\ref{eq:separated_power_law_non_log} and Equation~\ref{eq:quadratic_interaction_law}, we
observed that when the number of experts ($E$) remains fixed, these equations can be simplified to the first term in Equation~\ref{eq:L_D_N_sigma}. Motivated by these insights, we introduce a unified scaling law for both Dense Models and MoE Models, represented by Equation~\ref{eq:loss_N_D_E}. Specifically, since the decrease in performance is observed only when the number of experts ($E$) is large, we adopt the simplified one (Equation~\ref{eq:separated_power_law}) for small $E$ value (below 100). Besides, previous work \citep{bi2024deepseek} suggests replacing the model size $P$ (number of parameters) with model scale $N$, which is the result of the FLOPs divided by the number of tokens in order to fit Equation~\ref{eq:L_D_N_sigma} more accurately. Finally, we get Equation~\ref{eq:loss_N_D_E}:
\begin{equation}
    \hat{L}(N,D,E) = \frac{A}{N^\alpha E^\gamma} + \frac{B}{D^\beta} +\sigma
    \label{eq:loss_N_D_E}
\end{equation}
\[
\text{s.t.} \quad \text{FLOPs}(N, D) = C
\]
\noindent where $L$ is the training loss, $D$ is the number of training tokens, and $N$ is the model scale, which is the non-embedding FLOPs ($C$) divided by $D$. $E$ is the number of experts and we suggest $E$ is smaller than 100. $\sigma$ roughly estimates the natural noise of the dataset, representing the minimum achievable training loss. $A$, $B$, $\alpha$, $\beta$ and $\gamma$ are coefficients.

In order to validate Equation~\ref{eq:loss_N_D_E}, in our experiment, we fitted the training data for both 200M and 700M MoE Models (both with Eight Experts) to a curve. We then used this curve to predict the training loss scaling behavior of a 1.5B MoE model. The results, shown in Figure~\ref{fig:training_loss_extrapolation_1500m}, demonstrate that the formula we proposed based on previous work is applicable to MoE Models when the number of experts is not large. It could be observed that the reduction in benefits from increasing $E$ is minimal and the scaling equation stands within our experiment scope. Therefore, we adopted the simplified version. This formula suggests that the Equation~\ref{eq:loss_N_D_E} could equal to Equation~\ref{eq:L_D_N_sigma}, given a fixed number of experts, for MoE Models. This consistency could help us to compare the computing resource allocation strategy scaling for Dense Model and MoE Models (when $E$ is given).

\subsection{Estimating Optimal Resource Allocation Strategy Scaling}

After fitting the training loss ($L$) as a function of the number of tokens ($D$), the model scale ($N$), and the number of experts ($E$), we proceed to derive the optimal computing resource allocation strategy for model scale and the number of training tokens given a fixed compute budget.

The objective can be defined as follows: given a fixed number of compute budget, how to estimate the optimal resource allocation strategy for model scale and the number of training tokens to minimize the training loss.
\begin{equation}
    D_{\text{opt}}(C), N_{\text{opt}}(C) = \argmin_{D, N \text{s.t.} C = ND}  L(N, D, E)
\label{eq:compute_budget_allocation}
\end{equation}
We take the differentiation of Equation \ref{eq:loss_N_D_E} with respect to $C=DN$, we derive the optimal values of $D$ and $N$ for a given compute budget:
\begin{equation}
    \hat{D}_{opt}(C) = k_D \cdot C^{\alpha_D}
\label{eq:D_opt}
\end{equation}
\noindent where $k_D = (\frac{\alpha A}{\beta B E^\gamma})^{-{\frac{1}{\alpha + \beta}}}$ and $\alpha_D = \frac{\alpha}{\alpha + \beta}$
\begin{equation}
    \hat{N}_{opt}(C) = k_N \cdot C^{\alpha_N}
\label{eq:N_opt}
\end{equation}
\noindent where $k_N = (\frac{\alpha A}{\beta B E^\gamma})^{{\frac{1}{\alpha + \beta}}}$ and $\alpha_N = \frac{\beta}{\alpha + \beta}$. Equation ~\ref{eq:D_opt} and ~\ref{eq:N_opt} indicate how the model scale (compute budget per token $N$) and the optimal number of training tokens $D$ scale with the overall compute budget $C$. The most crucial part of this process is to find the scaling exponent of the model scale and tokens number with reference to the compute budget. Then we use the empirical fitting results obtained previously to calculate these two exponents.

\begin{table}[htbp]
\centering 
\fontsize{7.5}{7.5}\selectfont
\caption{Coefficients of optimal model and data scaling allocation for different model architectures.}
\label{tab1}
\renewcommand{\arraystretch}{2}
\begin{tabular}{ccccc}
\toprule
Model Architecture & $\alpha_D$ & $\alpha_N$ \\
\midrule
OpenAI (OpenWebText2) & 0.27 & 0.73  \\
Chinchilla (MassiveText) & 0.51 & 0.49\\ 
DeepSeek (OpenWebText2) & 0.422 & 0.578 \\
\midrule
Dense Model & 0.493 & 0.507 \\
MoE Model & 0.410 & 0.590\\ 
\bottomrule
\label{tab:optimal_allocation}
\end{tabular}
\end{table}

From the data shown in Table ~\ref{tab:optimal_allocation}, we observe that the exponent for the optimal model scale ($N$) in Mixture of Experts (MoE) models is larger, while the exponent for the optimal number of training tokens ($D$) is smaller, compared to their Dense Model counterparts. This suggests that MoE Models benefit more from increasing model size relative to the number of training tokens.

This finding helps explain why MoE Models can outperform larger Dense Models. The higher utilization of training data by MoE Models allows them to leverage their diverse sub-networks more effectively, capturing a broader range of features and patterns. Moreover, this suggests that it is more advantageous to allocate a larger computing budget to MoE Models compared to Dense Models.

\section{Optimal Hyperparameters Scaling}
\subsection{{Estimating Optimal Batch Size}}
\begin{figure*}[ht]
	\centering
	\vspace{-0.15in}
	\begin{minipage}{1\linewidth}
            \subfigure[]{
			\label{fig:heatmap_700m_128}			\includegraphics[width=0.50\linewidth,height=2.1in]{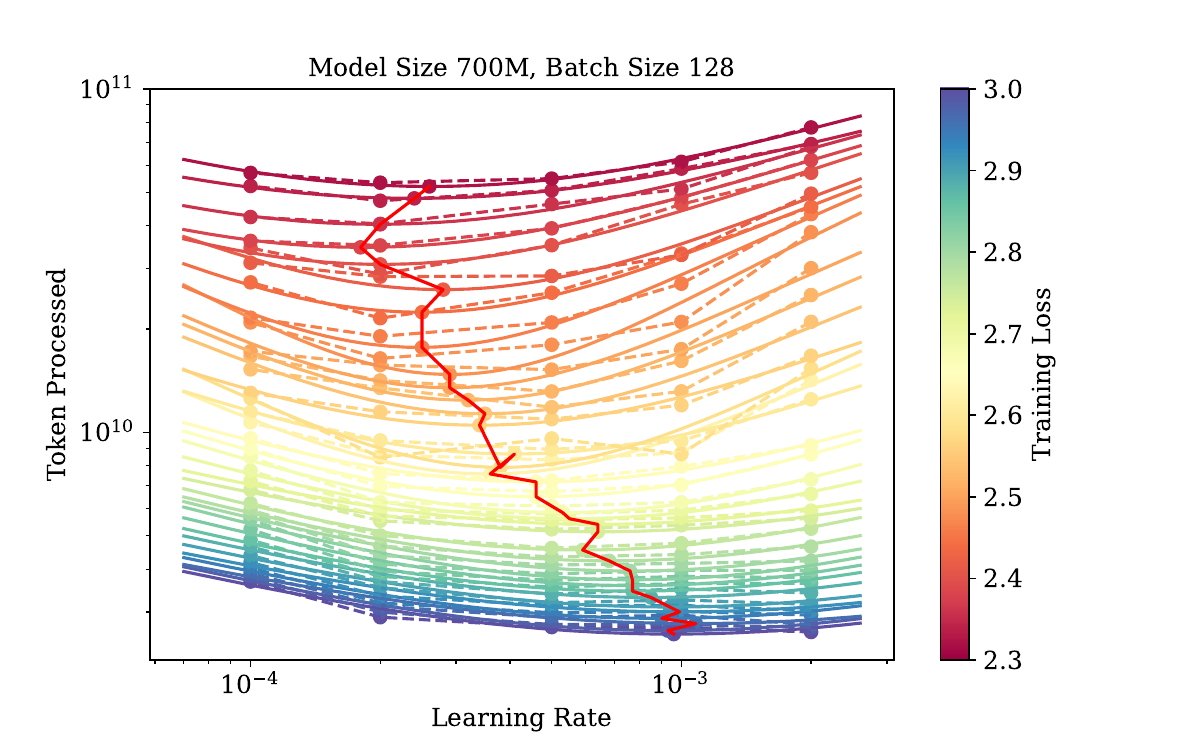}	
		}\noindent
            \subfigure[]{
			\label{fig:heatmap_1500m_128}			\includegraphics[width=0.50\linewidth,height=2.1in]{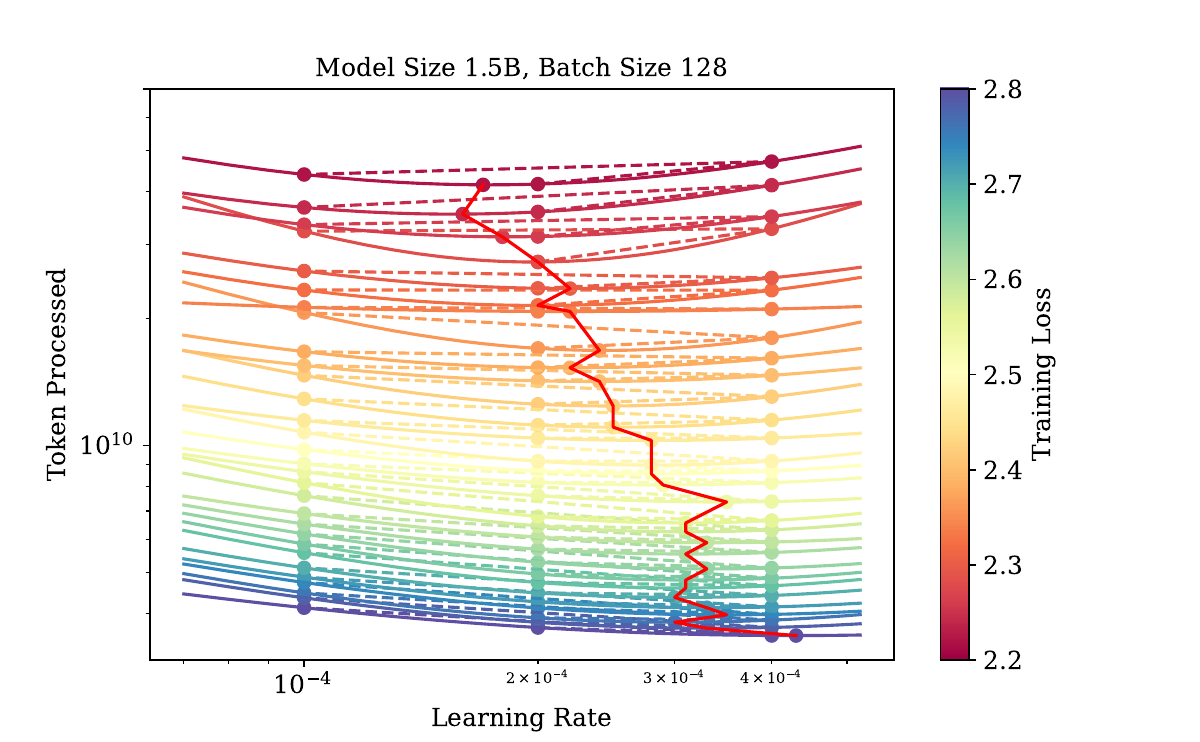}	
		}\noindent
	\end{minipage}
	\vspace{-0.15in}	
	\caption{This diagram displays a heatmap showing the distribution of training loss in relation to optimal learning rates and training token quantities, with fitted curves representing different training loss values. A vertical red line connects the minimum points of each curve. (a) Training loss vs. optimal learning rate when the MoE model size is 700M and the batch size is 128 (the number of sequences of length 8192). (b) Training loss vs. optimal learning rate when the MoE model size is 1.5B, with the batch size being 128 (the number of sequences of length 8192).}
	\vspace{-0.in}
	\label{fig:heatmap_lr_loss}
\end{figure*}

Batch size is a crucial hyperparameter for the training process. Previous works \citep{mccandlish2018empirical, kaplan2020scaling} have investigated the scaling law between optimal batch size and loss for Dense Models. In this experiment, we conduct the analysis of both models.

To start, according to previous work \citep{mccandlish2018empirical}, let $\Delta L_{\text{opt}}(B)$ denote the optimal improvement in the loss function when using a batch size $B$ with the optimal step size $\epsilon_{\text{opt}}(B)$. It takes into account the noise introduced by the gradient estimation process. Then $\Delta L_{\text{opt}}(B)$ could be shown in Equation~\ref{delta_b_b_noise_b}.
\begin{equation}
    \Delta L_{\text{opt}}(B) = \Delta L_{\text{max}} \left(1 + \frac{B_{\text{noise}}}{B}\right)^{-1}
\label{delta_b_b_noise_b}
\end{equation}
\noindent where $\Delta L_{\text{max}}$ is the maximum possible improvement in the loss function when the true gradient is used without noise. Here, the noise scale \( B_{\text{noise}} \)  measures the scale of the noise in the gradient estimates relative to the true gradient. It helps quantify how much the noise affects the gradient estimation process. The noise scale could be defined as:

\begin{equation}
B_{\text{noise}} = \frac{\text{tr} (H\Sigma)}{G^T H G} \quad 
\label{eq:b_opt_def}
\end{equation}
\noindent where $G$ denotes the true gradient and $H$ the true Hessian at parameter values $\theta$. 
$\Sigma$ represents covariance matrix, and $tr(\cdot)$ represents the trace. A higher noise scale ($B_{\text{noise}}$) implies a larger optimal batch size for reducing the variance in the gradient estimate. 

For training efficiency, the relationship between training steps \( S \) and training examples \( E \) derived from the SGD optimizer could be expressed as:
\begin{equation}
(\frac{S}{S_{\text{min}}} - 1)(\frac{E}{E_{\text{min}}} - 1) = 1 \quad
\label{eq:training_speed_data_efficiency}
\end{equation}
\noindent where \( S_{\text{min}} \) and \( E_{\text{min}} \) are the minimum training steps and training examples needed to achieve a specific performance. Finally, from the empirical and theoretical verification \citep{mccandlish2018empirical,kaplan2020scaling, hu2024minicpm}, the optimal batch size at a specific training loss could be approximated using $B_{\text{opt}} \approx B_{\text{noise}}$, then we get Equation~\ref{eq:loss_opt_bs}.
\begin{equation}
   B_{\text{noise}} \approx B_{\text{opt}} = \frac{E_{\text{min}}}{S_{\text{min}}} \approx \frac{\lambda_{B}}{L^{\alpha_{B}}}
\label{eq:loss_opt_bs}
\end{equation}
\noindent where $\lambda_{B}$ and $\alpha_{B}$ are both coefficients. $B_{opt}$ is the optimal batch size given a noisy gradient and $L$ is the loss value.

Equation ~\ref{eq:loss_opt_bs} indicates that \( B_{\text{opt}} \) serves as the balance point between training speed and data efficiency. It represents the optimal trade-off between training speed and data efficiency. Furthermore, it indicates that as training progresses and the loss decreases, \( B_{\text{opt}} \) gradually becomes larger, indicating that larger batch size is required to maintain the balance between training speed and data efficiency as the model converges. 

Then we investigate the Equation~\ref{eq:loss_opt_bs} for both Dense Models and MoE Models (8 experts). Following the procedure in \citep{hu2024minicpm}, we first generate contour lines representing configurations with an equivalent number of training tokens (Figure ~\ref{fig:heatmap_bs_loss}) for both Dense Models and MoE Models. From each contour line, we identify the points that exhibited the minimum training loss along with their corresponding batch size and learning rate. Subsequently, these optimal points are plotted in Figure~\ref{fig:optimal_bs_loss_dense} and Figure~\ref{fig:optimal_bs_loss_MoE} to illustrate the relationship between training loss and batch size. We could observe that Dense Model has a larger optimal batch size exponent, which means for the same task and training loss value, the noise scale for Dense Model is larger than that of MoE Models.

Through the theoretical analysis (Equation~\ref{eq:b_opt_def} and ~\ref{eq:loss_opt_bs}), the noise scale $B_{noise}$ is related to the variance of the gradient estimates and $B_{opt}$ could be approximate to the noise scale. The observation that MoE Models have smaller optimal batch sizes for the same loss suggests that MoE Models can achieve the same optimization efficiency with less data, indicating a smaller noise scale. However, for Dense Models, larger optimal batch sizes for the same loss suggest that dense models need more data to achieve the same optimization efficiency, indicating a larger noise scale. This means that the gradient estimates are noisier in dense models for a given batch size.

\subsection{Estimating Optimal Learning Rate}

\begin{figure*}[th]
	\centering
	\vspace{-0.15in}
	\begin{minipage}{1\linewidth}
            \subfigure[]{
			\label{fig:optimal_lr_loss_dense}			\includegraphics[width=0.48\linewidth,height=2.1in]{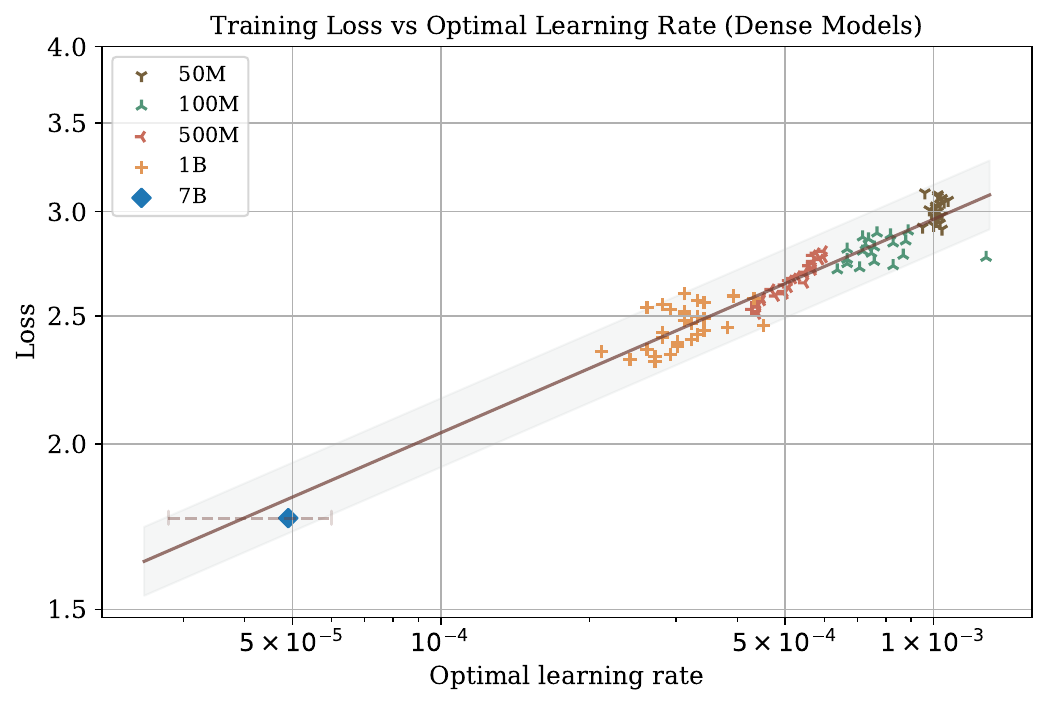}	
		}\noindent
            \subfigure[]{
			\label{fig:optimal_lr_loss_MoE}			\includegraphics[width=0.48\linewidth,height=2.1in]{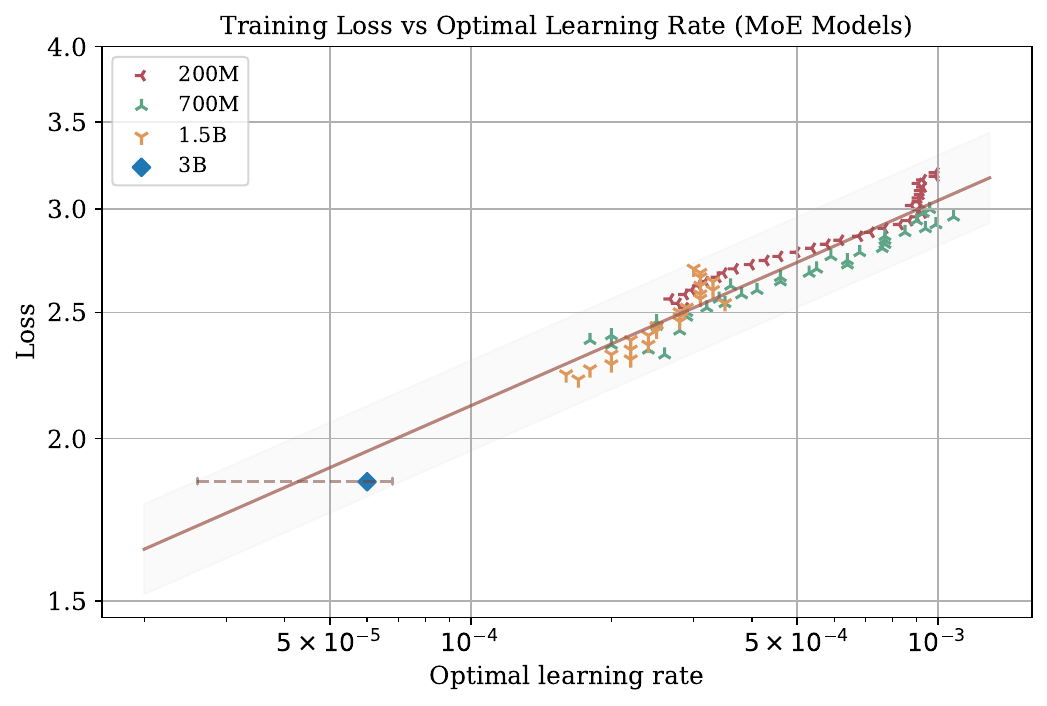}	
		}\noindent
	\end{minipage}
	\vspace{-0.15in}	
	\caption{We plot the optimal learning rate values together with the corresponding training loss values across different model sizes for both Dense Models and MoE Models. In log scale diagrams, (a) demonstrates the log-log relationship of training loss vs. optimal learning rate for Dense Models. (b) demonstrates the log-log relationship of training loss vs. optimal learning rate for MoE Models. This indicates that the power-law relationship remains consistent not only across model sizes but also across model architectures. The total overlap of the comparative performance interval is about 76.2\%.} 
	\vspace{-0.15in}
	\label{fig:optimal_lr_loss}
\end{figure*}

In previous work \citep{mccandlish2018empirical}, the scaling relationship between optimal learning rate and optimal batch size when using SGD optimizer are illustrated as Equation ~\ref{eq:sgd_opt_lr}. 
\begin{equation}
    \epsilon_{\text{opt}}(B) = \epsilon_{\text{max}} \left( 1 + \frac{B_{\text{noise}}}{B} \right)^{-1} \quad 
\label{eq:sgd_opt_lr}
\end{equation}
\noindent where $\epsilon_{\text{opt}}(B)$ represents the optimal step size that minimizes the expected loss from the noisy gradient. And $\epsilon_{max}$ represents the optimal step size that minimizes the loss function when using the noiseless true gradient $G$ to update the parameters. It is defined as $\epsilon_{max} = \frac{|G|^2}{G^T HG}$. This relationship (Equation~\ref{eq:sgd_opt_lr}) shows that when $B$ is relatively small, this Equation could be reduced as a nearly linear scaling \citep{granziol2022learning, goyal2017accurate}. When the batch size is fixed, with optimal learning rate decreases with the increasing of noise scale.
\[
\epsilon_{\text{opt}}(B) = \epsilon_{\text{max}} \left( \frac{B}{B_{\text{noise}}} \right)
\]
Recent research by \citet{li2024surge} and \citet{granziol2022learning} reveals an interesting trend in the optimal learning rate for the Adam Optimizer concerning the optimal batch size $B_{\text{opt}}$ or noise scale $B_{\text{noise}}$. It shows a non-monotonic behavior, indicating that as the noise scale $B_{\text{noise}}$ increases, the optimal learning rate initially follows a nearly square root scaling pattern, dominated by the second term, before decreasing as the first term gains dominance. The transition point occurs when these two terms reach a balance, a critical juncture determined by the specific values of $B_{\text{noise}}$ and $B$.

\begin{equation}
    \epsilon_{\text{opt}}(B) = \frac{2 \epsilon_{\text{max}}}{  \sqrt{\frac{B_{\text{noise}}}{B}} + \sqrt{\frac{B}{B_{\text{noise}}}} \quad}
\label{eq:adam_opt_lr}
\end{equation}
In order to verify this, in this experiment (shown in Figure~\ref{fig:heatmap_lr_loss}), we plot contour lines that represent configurations with an equivalent number of training tokens. For each contour line, we identify the points that achieved the minimum training loss, recording their corresponding learning rate. Then Figure~\ref{fig:optimal_lr_loss} illustrates the relationship between training loss and optimal learning rate in Equation~\ref{eq:loss_lr}. 
\begin{equation}
    \epsilon_{opt} \approx \frac{\lambda_{\epsilon}}{L^{\alpha_{\epsilon}}}
\label{eq:loss_lr}
\end{equation}
\noindent where $\lambda_{\epsilon}$ and $\alpha_{\epsilon}$ are both coefficients. $\epsilon_{opt}$ is the optimal learning rate given a noisy gradient and $L$ is the loss value.

From the observation, MoE Models are likely to have a larger optimal learning rate compared to Dense Models when assuming the same loss value. Firstly, MoE Models tend to have smaller noise scale for the same loss value compared to Dense Models. A smaller noise scale indicates that the gradient estimates are less noisy, allowing for more accurate updates with smaller batch sizes. This efficiency with smaller batches also translates into requiring larger learning rate for effective optimization. Secondly, the optimal learning rate is roughly inversely proportional to the noise scale when the first term dominates for Equation~\ref{eq:adam_opt_lr}. In conclusion, MoE Models are generally more efficient with smaller batch sizes and larger learning rates. It is very likely the model architecture that makes MoE models more efficient and more able to handle complex data.

\section{Generalization of Scaling Law}

\begin{figure}[t]
	\centering
	\vspace{-0.15in}
	\label{fig:test_loss_compute_budget_dense_moe}			              
        \includegraphics[width=0.99\linewidth,height=2.1in]{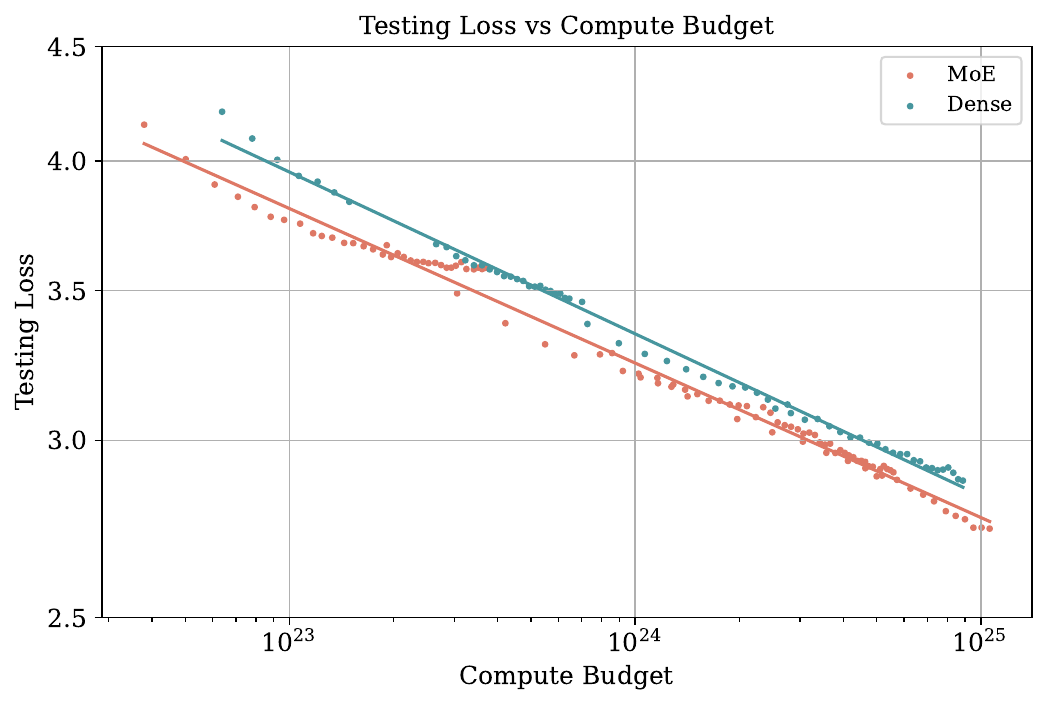}	
	\vspace{-0.1in}	
	\caption{We explore the generalization ability for both Dense Models and MoE Models. We highlight the stable power-law relationship interval across different model sizes for testing loss and the compute budget. We observe that with a comparable compute budget, MoE Models could achieve better testing loss.}
	\vspace{-0.15in}
	\label{fig:train_test_loss}
\end{figure}

\begin{table}[h]
\centering 
\fontsize{6.5}{7.5}\selectfont
\caption{Details about the performance of our Dense and MoE Models involved in all experiments.}
\label{tab1}
\renewcommand{\arraystretch}{2.1}
\begin{tabular}{cccccccccc}
\toprule
Model Size & TriviaQA & MATH & MMLU & CMMLU & MATH401 \\
\midrule
Dense-7B & 54.67 & 5.26 & 28.73 & 25.03 & 49.13 \\
Dense-1B & 20.56 & 1.48 & 22.42 & 21.38 & 5.99 \\
Dense-500M & 17.86 & 1.18 & 14.65 & 20.02 & 4.99 \\
Dense-100M & 14.28 & 0.66 & 14.68 & 18.71 & 3.74 \\
Dense-50M & 13.47 & 0.96 & 11.17 & 22.84 & 1.75\\
\midrule
MoE-3B & 56.09 & 6.74 & 29.18 & 25.46 & 40.12 \\
MoE-1.5B & 26.25 & 4.24 & 25.26 & 19.81 & 7.02 \\
MoE-700M & 18.89 & 1.80 & 15.51 & 18.04 & 4.68 \\
MoE-200M & 14.36 & 0.96 & 14.40 & 14.60 & 2.77 \\
\bottomrule
\end{tabular}
\label{tab:perf_dense_moe}
\end{table}

We predict that MoE Models could have better generalization ability compared to their Dense counterparts. Firstly, MoE Models consist of multiple expert sub-networks that specialize in different aspects of the data, which enables the model to capture a broader range of features and patterns and then leads to better generalization across various tasks and datasets. Additionally, the ensemble nature of MoE Models, where each expert contributes to the final prediction, reduces overfitting and improves robustness by combining predictions from multiple models. Finally, the gating mechanisms in MoE Models control the contribution of each expert to the final prediction, acting as a form of regularization. This regularization helps prevent overfitting by focusing on the most relevant experts for each input, resulting in better generalization.

To investigate the generalization performance of the scaling law for both MoE Models (Eight Experts) and Dense Models, we explore and compare the relationship of training loss with testing loss and compute budget for both Dense Models and MoE Models (shown in Figure~\ref{fig:train_test_loss}). We highlight the stable power-law relationship interval across different model sizes, illustrating the correlation between testing loss and compute budget. Notably, MoE Models consistently exhibit a smaller testing loss for a given compute budget, indicating their superior generalization performance.

We also explore the performance of the Dense and MoE Models on different testing sets in Table~\ref{tab:perf_dense_moe}. It clearly shows that MoE Models could outperform a Dense Model with comparative model size. The consistent trend in performance across different datasets underscores the transferability and reliability of the scaling law observed in our study. This finding suggests that the performance improvements achieved by MoE Models are not limited to specific datasets or conditions but hold true across diverse testing sets, indicating robustness and generalizability in real-world applications. 

\section{Conclusion}
In this paper, we investigate the transfer of traditional scaling laws from Dense Models to Mixture of Experts (MoE) Models. Our investigation confirms that the power-law relationship extends to MoE Models regarding the consistency and transferability of scaling strategies, including resource allocation strategy, optimal batch size / learning rate scaling, and so on. This observation indicates that the fundamental principles of training dynamics and the behavior of scaling rules are similar for both Models. This means existing knowledge and practices for optimizing Dense Models can be easily adapted to MoE Models, potentially reducing the experimental burden of finding the best hyperparameters. Besides, we find that MoE Models demonstrate approximately a 16.37\% improvement in data utilization efficiency compared to Dense Models with a fixed compute budget. Thus we suggest prioritizing increasing model scale over other factors when training MoE Models, highlighting their greater data efficiency. Finally, we use both theoretical and empirical analysis to reveal that during the training process, MoE Models exhibit a lower gradient noise scale when using the Adam Optimizer. At the same training loss value, MoE Models can achieve stable training with smaller batch sizes and larger learning rates, potentially speeding up the training process and improving convergence. It shows that MoE Models can make more efficient use of training data and computational resources, extracting more information per training token, leading to faster training times and better utilization of available data. These results offer valuable insights for refining training and deployment strategies for MoE Models.

\section*{Limitation}

The experiments were constrained by the available computational resources. Firstly, we have not yet explored scenarios with more than 100 experts in our experiments. It has been observed that as the model size increases, the marginal benefit from increasing the number of experts tends to decrease. Moreover, although we tested our model on several benchmark datasets, they don't cover all domains. To ensure robust evaluation, we plan to validate the model on more tasks and domains and investigate the scaling relationships between different metrics. Additionally, as the number of training tokens exceeded 100 billion, we observed signs of overtraining for both models, indicating diminishing returns in performance improvements with additional training data. These areas remain open for future research.

\section*{Acknowledgement}
Jingang Wang is funded by Beijing Nova Program (Grant NO. 20220484098).

\bibliography{custom}

\begin{thebibliography}{47}
\providecommand{\natexlab}[1]{#1}

\bibitem[{Achiam et~al.(2023)Achiam, Adler, Agarwal, Ahmad, Akkaya, Aleman,
  Almeida, Altenschmidt, Altman, Anadkat et~al.}]{achiam2023gpt}
Josh Achiam, Steven Adler, Sandhini Agarwal, Lama Ahmad, Ilge Akkaya,
  Florencia~Leoni Aleman, Diogo Almeida, Janko Altenschmidt, Sam Altman,
  Shyamal Anadkat, et~al. 2023.
\newblock Gpt-4 technical report.
\newblock \emph{arXiv preprint arXiv:2303.08774}.

\bibitem[{Bahri et~al.(2021)Bahri, Dyer, Kaplan, Lee, and
  Sharma}]{bahri2021explaining}
Yasaman Bahri, Ethan Dyer, Jared Kaplan, Jaehoon Lee, and Utkarsh Sharma. 2021.
\newblock Explaining neural scaling laws.
\newblock \emph{arXiv preprint arXiv:2102.06701}.

\bibitem[{Bi et~al.(2024)Bi, Chen, Chen, Chen, Dai, Deng, Ding, Dong, Du, Fu
  et~al.}]{bi2024deepseek}
Xiao Bi, Deli Chen, Guanting Chen, Shanhuang Chen, Damai Dai, Chengqi Deng,
  Honghui Ding, Kai Dong, Qiushi Du, Zhe Fu, et~al. 2024.
\newblock Deepseek llm: Scaling open-source language models with longtermism.
\newblock \emph{arXiv preprint arXiv:2401.02954}.

\bibitem[{Brown et~al.(2020)Brown, Mann, Ryder, Subbiah, Kaplan, Dhariwal,
  Neelakantan, Shyam, Sastry, Askell et~al.}]{brown2020language}
Tom Brown, Benjamin Mann, Nick Ryder, Melanie Subbiah, Jared~D Kaplan, Prafulla
  Dhariwal, Arvind Neelakantan, Pranav Shyam, Girish Sastry, Amanda Askell,
  et~al. 2020.
\newblock Language models are few-shot learners.
\newblock \emph{Advances in neural information processing systems},
  33:1877--1901.

\bibitem[{Chen et~al.(2021{\natexlab{a}})Chen, Ge, Zhan, Huang, and
  Wang}]{chen2021pareto}
Zhengyu Chen, Jixie Ge, Heshen Zhan, Siteng Huang, and Donglin Wang.
  2021{\natexlab{a}}.
\newblock Pareto self-supervised training for few-shot learning.
\newblock In \emph{Proceedings of the IEEE/CVF Conference on Computer Vision
  and Pattern Recognition}, pages 13663--13672.

\bibitem[{Chen and Wang(2021)}]{chen2021multi}
Zhengyu Chen and Donglin Wang. 2021.
\newblock Multi-initialization meta-learning with domain adaptation.
\newblock In \emph{ICASSP 2021-2021 IEEE International Conference on Acoustics,
  Speech and Signal Processing (ICASSP)}, pages 1390--1394. IEEE.

\bibitem[{Chen et~al.(2022)Chen, Xiao, and Kuang}]{chen2022ba}
Zhengyu Chen, Teng Xiao, and Kun Kuang. 2022.
\newblock Ba-gnn: On learning bias-aware graph neural network.
\newblock In \emph{2022 IEEE 38th International Conference on Data Engineering
  (ICDE)}, pages 3012--3024. IEEE.

\bibitem[{Chen et~al.(2024{\natexlab{a}})Chen, Xiao, Kuang, Lv, Zhang, Yang,
  Lu, Yang, and Wu}]{chen2024learning}
Zhengyu Chen, Teng Xiao, Kun Kuang, Zheqi Lv, Min Zhang, Jinluan Yang,
  Chengqiang Lu, Hongxia Yang, and Fei Wu. 2024{\natexlab{a}}.
\newblock Learning to reweight for generalizable graph neural network.
\newblock In \emph{Proceedings of the AAAI Conference on Artificial
  Intelligence}, volume~38, pages 8320--8328.

\bibitem[{Chen et~al.(2024{\natexlab{b}})Chen, Xiao, Wang, and
  Zhang}]{chen2024pareto}
Zhengyu Chen, Teng Xiao, Donglin Wang, and Min Zhang. 2024{\natexlab{b}}.
\newblock Pareto graph self-supervised learning.
\newblock In \emph{ICASSP 2024-2024 IEEE International Conference on Acoustics,
  Speech and Signal Processing (ICASSP)}, pages 6630--6634. IEEE.

\bibitem[{Chen et~al.(2021{\natexlab{b}})Chen, Xu, and Wang}]{chen2021deep}
Zhengyu Chen, Ziqing Xu, and Donglin Wang. 2021{\natexlab{b}}.
\newblock Deep transfer tensor decomposition with orthogonal constraint for
  recommender systems.
\newblock In \emph{Proceedings of the AAAI Conference on Artificial
  Intelligence}, volume~35, pages 4010--4018.

\bibitem[{Clark et~al.(2022)Clark, De~Las~Casas, Guy, Mensch, Paganini,
  Hoffmann, Damoc, Hechtman, Cai, Borgeaud, Van Den~Driessche, Rutherford,
  Hennigan, Johnson, Cassirer, Jones, Buchatskaya, Budden, Sifre, Osindero,
  Vinyals, Ranzato, Rae, Elsen, Kavukcuoglu, and Simonyan}]{pmlr-v162-clark22a}
Aidan Clark, Diego De~Las~Casas, Aurelia Guy, Arthur Mensch, Michela Paganini,
  Jordan Hoffmann, Bogdan Damoc, Blake Hechtman, Trevor Cai, Sebastian
  Borgeaud, George~Bm Van Den~Driessche, Eliza Rutherford, Tom Hennigan,
  Matthew~J Johnson, Albin Cassirer, Chris Jones, Elena Buchatskaya, David
  Budden, Laurent Sifre, Simon Osindero, Oriol Vinyals, Marc'Aurelio Ranzato,
  Jack Rae, Erich Elsen, Koray Kavukcuoglu, and Karen Simonyan. 2022.
\newblock \href {https://proceedings.mlr.press/v162/clark22a.html} {Unified
  scaling laws for routed language models}.
\newblock In \emph{Proceedings of the 39th International Conference on Machine
  Learning}, volume 162 of \emph{Proceedings of Machine Learning Research},
  pages 4057--4086. PMLR.

\bibitem[{Du et~al.(2022)Du, Huang, Dai, Tong, Lepikhin, Xu, Krikun, Zhou, Yu,
  Firat et~al.}]{du2022glam}
Nan Du, Yanping Huang, Andrew~M Dai, Simon Tong, Dmitry Lepikhin, Yuanzhong Xu,
  Maxim Krikun, Yanqi Zhou, Adams~Wei Yu, Orhan Firat, et~al. 2022.
\newblock Glam: Efficient scaling of language models with mixture-of-experts.
\newblock In \emph{International Conference on Machine Learning}, pages
  5547--5569. PMLR.

\bibitem[{Fedus et~al.(2022)Fedus, Zoph, and Shazeer}]{fedus2022switch}
William Fedus, Barret Zoph, and Noam Shazeer. 2022.
\newblock Switch transformers: Scaling to trillion parameter models with simple
  and efficient sparsity.
\newblock \emph{Journal of Machine Learning Research}, 23(120):1--39.

\bibitem[{Gadre et~al.(2024)Gadre, Smyrnis, Shankar, Gururangan, Wortsman,
  Shao, Mercat, Fang, Li, Keh et~al.}]{gadre2024language}
Samir~Yitzhak Gadre, Georgios Smyrnis, Vaishaal Shankar, Suchin Gururangan,
  Mitchell Wortsman, Rulin Shao, Jean Mercat, Alex Fang, Jeffrey Li, Sedrick
  Keh, et~al. 2024.
\newblock Language models scale reliably with over-training and on downstream
  tasks.
\newblock \emph{arXiv preprint arXiv:2403.08540}.

\bibitem[{{Gao} et~al.(2020){Gao}, {Biderman}, {Black}, {Golding}, {Hoppe},
  {Foster}, {Phang}, {He}, {Thite}, {Nabeshima}, {Presser}, and
  {Leahy}}]{2021arXiv210100027G}
Leo {Gao}, Stella {Biderman}, Sid {Black}, Laurence {Golding}, Travis {Hoppe},
  Charles {Foster}, Jason {Phang}, Horace {He}, Anish {Thite}, Noa {Nabeshima},
  Shawn {Presser}, and Connor {Leahy}. 2020.
\newblock \href {https://doi.org/10.48550/arXiv.2101.00027} {{The Pile: An
  800GB Dataset of Diverse Text for Language Modeling}}.
\newblock \emph{arXiv e-prints}, arXiv:2101.00027.

\bibitem[{Garcia and Firat(2022)}]{garcia2022using}
Xavier Garcia and Orhan Firat. 2022.
\newblock Using natural language prompts for machine translation.
\newblock \emph{arXiv preprint arXiv:2202.11822}.

\bibitem[{Goyal et~al.(2017)Goyal, Doll{\'a}r, Girshick, Noordhuis, Wesolowski,
  Kyrola, Tulloch, Jia, and He}]{goyal2017accurate}
Priya Goyal, Piotr Doll{\'a}r, Ross Girshick, Pieter Noordhuis, Lukasz
  Wesolowski, Aapo Kyrola, Andrew Tulloch, Yangqing Jia, and Kaiming He. 2017.
\newblock Accurate, large minibatch sgd: Training imagenet in 1 hour.
\newblock \emph{arXiv preprint arXiv:1706.02677}.

\bibitem[{Granziol et~al.(2022)Granziol, Zohren, and
  Roberts}]{granziol2022learning}
Diego Granziol, Stefan Zohren, and Stephen Roberts. 2022.
\newblock Learning rates as a function of batch size: A random matrix theory
  approach to neural network training.
\newblock \emph{Journal of Machine Learning Research}, 23(173):1--65.

\bibitem[{Hendy et~al.(2023)Hendy, Abdelrehim, Sharaf, Raunak, Gabr,
  Matsushita, Kim, Afify, and Awadalla}]{Hendy2023HowGA}
Amr Hendy, Mohamed~Gomaa Abdelrehim, Amr Sharaf, Vikas Raunak, Mohamed Gabr,
  Hitokazu Matsushita, Young~Jin Kim, Mohamed Afify, and Hany~Hassan Awadalla.
  2023.
\newblock \href {https://api.semanticscholar.org/CorpusID:257038384} {How good
  are gpt models at machine translation? a comprehensive evaluation}.
\newblock \emph{ArXiv}, abs/2302.09210.

\bibitem[{Hoffmann et~al.(2022)Hoffmann, Borgeaud, Mensch, Buchatskaya, Cai,
  Rutherford, de~Las~Casas, Hendricks, Welbl, Clark, Hennigan, Noland,
  Millican, van~den Driessche, Damoc, Guy, Osindero, Simonyan, Elsen, Rae,
  Vinyals, and Sifre}]{hoffmann2022training}
Jordan Hoffmann, Sebastian Borgeaud, Arthur Mensch, Elena Buchatskaya, Trevor
  Cai, Eliza Rutherford, Diego de~Las~Casas, Lisa~Anne Hendricks, Johannes
  Welbl, Aidan Clark, Tom Hennigan, Eric Noland, Katie Millican, George van~den
  Driessche, Bogdan Damoc, Aurelia Guy, Simon Osindero, Karen Simonyan, Erich
  Elsen, Jack~W. Rae, Oriol Vinyals, and Laurent Sifre. 2022.
\newblock \href {https://arxiv.org/abs/2203.15556} {Training compute-optimal
  large language models}.
\newblock \emph{Preprint}, arXiv:2203.15556.

\bibitem[{Horváth et~al.(2021)Horváth, Klein, Richtárik, and
  Archambeau}]{horváth2021hyperparameter}
Samuel Horváth, Aaron Klein, Peter Richtárik, and Cédric Archambeau. 2021.
\newblock \href {https://arxiv.org/abs/2102.12810} {Hyperparameter transfer
  learning with adaptive complexity}.
\newblock \emph{Preprint}, arXiv:2102.12810.

\bibitem[{Hu et~al.(2024)Hu, Tu, Han, He, Cui, Long, Zheng, Fang, Huang, Zhao
  et~al.}]{hu2024minicpm}
Shengding Hu, Yuge Tu, Xu~Han, Chaoqun He, Ganqu Cui, Xiang Long, Zhi Zheng,
  Yewei Fang, Yuxiang Huang, Weilin Zhao, et~al. 2024.
\newblock Minicpm: Unveiling the potential of small language models with
  scalable training strategies.
\newblock \emph{arXiv preprint arXiv:2404.06395}.

\bibitem[{Huang and Chang(2022)}]{huang2022towards}
Jie Huang and Kevin Chen-Chuan Chang. 2022.
\newblock Towards reasoning in large language models: A survey.
\newblock \emph{arXiv preprint arXiv:2212.10403}.

\bibitem[{Jiang et~al.(2023)Jiang, Sablayrolles, Mensch, Bamford, Chaplot,
  Casas, Bressand, Lengyel, Lample, Saulnier et~al.}]{jiang2023mistral}
Albert~Q Jiang, Alexandre Sablayrolles, Arthur Mensch, Chris Bamford,
  Devendra~Singh Chaplot, Diego de~las Casas, Florian Bressand, Gianna Lengyel,
  Guillaume Lample, Lucile Saulnier, et~al. 2023.
\newblock Mistral 7b.
\newblock \emph{arXiv preprint arXiv:2310.06825}.

\bibitem[{Jiang et~al.(2024)Jiang, Sablayrolles, Roux, Mensch, Savary, Bamford,
  Chaplot, Casas, Hanna, Bressand et~al.}]{jiang2024mixtral}
Albert~Q Jiang, Alexandre Sablayrolles, Antoine Roux, Arthur Mensch, Blanche
  Savary, Chris Bamford, Devendra~Singh Chaplot, Diego de~las Casas, Emma~Bou
  Hanna, Florian Bressand, et~al. 2024.
\newblock Mixtral of experts.
\newblock \emph{arXiv preprint arXiv:2401.04088}.

\bibitem[{Kaplan et~al.(2020)Kaplan, McCandlish, Henighan, Brown, Chess, Child,
  Gray, Radford, Wu, and Amodei}]{kaplan2020scaling}
Jared Kaplan, Sam McCandlish, Tom Henighan, Tom~B. Brown, Benjamin Chess, Rewon
  Child, Scott Gray, Alec Radford, Jeffrey Wu, and Dario Amodei. 2020.
\newblock \href {https://arxiv.org/abs/2001.08361} {Scaling laws for neural
  language models}.
\newblock \emph{Preprint}, arXiv:2001.08361.

\bibitem[{Lepikhin et~al.(2020)Lepikhin, Lee, Xu, Chen, Firat, Huang, Krikun,
  Shazeer, and Chen}]{lepikhin2020gshard}
Dmitry Lepikhin, HyoukJoong Lee, Yuanzhong Xu, Dehao Chen, Orhan Firat, Yanping
  Huang, Maxim Krikun, Noam Shazeer, and Zhifeng Chen. 2020.
\newblock Gshard: Scaling giant models with conditional computation and
  automatic sharding.
\newblock \emph{arXiv preprint arXiv:2006.16668}.

\bibitem[{Li et~al.(2024)Li, Zhao, Zhang, Sun, Wu, Jiao, Wang, Liu, Fang, Xue
  et~al.}]{li2024surge}
Shuaipeng Li, Penghao Zhao, Hailin Zhang, Xingwu Sun, Hao Wu, Dian Jiao, Weiyan
  Wang, Chengjun Liu, Zheng Fang, Jinbao Xue, et~al. 2024.
\newblock Surge phenomenon in optimal learning rate and batch size scaling.
\newblock \emph{arXiv preprint arXiv:2405.14578}.

\bibitem[{McCandlish et~al.(2018)McCandlish, Kaplan, Amodei, and
  Team}]{mccandlish2018empirical}
Sam McCandlish, Jared Kaplan, Dario Amodei, and OpenAI~Dota Team. 2018.
\newblock An empirical model of large-batch training.
\newblock \emph{arXiv preprint arXiv:1812.06162}.

\bibitem[{Muennighoff et~al.(2024)Muennighoff, Rush, Barak, Le~Scao, Tazi,
  Piktus, Pyysalo, Wolf, and Raffel}]{muennighoff2024scaling}
Niklas Muennighoff, Alexander Rush, Boaz Barak, Teven Le~Scao, Nouamane Tazi,
  Aleksandra Piktus, Sampo Pyysalo, Thomas Wolf, and Colin~A Raffel. 2024.
\newblock Scaling data-constrained language models.
\newblock \emph{Advances in Neural Information Processing Systems}, 36.

\bibitem[{Perrone et~al.(2018)Perrone, Jenatton, Seeger, and
  Archambeau}]{Perrone2018ScalableHT}
Valerio Perrone, Rodolphe Jenatton, Matthias~W. Seeger, and C.~Archambeau.
  2018.
\newblock \href {https://api.semanticscholar.org/CorpusID:54035096} {Scalable
  hyperparameter transfer learning}.
\newblock In \emph{Neural Information Processing Systems}.

\bibitem[{Radford et~al.(2019)Radford, Wu, Child, Luan, Amodei, and
  Sutskever}]{Radford2019LanguageMA}
Alec Radford, Jeff Wu, Rewon Child, David Luan, Dario Amodei, and Ilya
  Sutskever. 2019.
\newblock \href {https://api.semanticscholar.org/CorpusID:160025533} {Language
  models are unsupervised multitask learners}.

\bibitem[{Rae et~al.(2021)Rae, Borgeaud, Cai, Millican, Hoffmann, Song,
  Aslanides, Henderson, Ring, Young et~al.}]{rae2021scaling}
Jack~W Rae, Sebastian Borgeaud, Trevor Cai, Katie Millican, Jordan Hoffmann,
  Francis Song, John Aslanides, Sarah Henderson, Roman Ring, Susannah Young,
  et~al. 2021.
\newblock Scaling language models: Methods, analysis \& insights from training
  gopher.
\newblock \emph{arXiv preprint arXiv:2112.11446}.

\bibitem[{Rajbhandari et~al.(2022)Rajbhandari, Li, Yao, Zhang, Aminabadi, Awan,
  Rasley, and He}]{rajbhandari2022deepspeed}
Samyam Rajbhandari, Conglong Li, Zhewei Yao, Minjia Zhang, Reza~Yazdani
  Aminabadi, Ammar~Ahmad Awan, Jeff Rasley, and Yuxiong He. 2022.
\newblock Deepspeed-moe: Advancing mixture-of-experts inference and training to
  power next-generation ai scale.
\newblock In \emph{International conference on machine learning}, pages
  18332--18346. PMLR.

\bibitem[{Shazeer et~al.(2017)Shazeer, Mirhoseini, Maziarz, Davis, Le, Hinton,
  and Dean}]{shazeer2017outrageously}
Noam Shazeer, Azalia Mirhoseini, Krzysztof Maziarz, Andy Davis, Quoc Le,
  Geoffrey Hinton, and Jeff Dean. 2017.
\newblock Outrageously large neural networks: The sparsely-gated
  mixture-of-experts layer.
\newblock \emph{arXiv preprint arXiv:1701.06538}.

\bibitem[{Shen et~al.(2024)Shen, Guo, Cai, and Qin}]{shen2024jetmoe}
Yikang Shen, Zhen Guo, Tianle Cai, and Zengyi Qin. 2024.
\newblock Jetmoe: Reaching llama2 performance with 0.1 m dollars.
\newblock \emph{arXiv preprint arXiv:2404.07413}.

\bibitem[{Team et~al.(2023)Team, Anil, Borgeaud, Wu, Alayrac, Yu, Soricut,
  Schalkwyk, Dai, Hauth et~al.}]{team2023gemini}
Gemini Team, Rohan Anil, Sebastian Borgeaud, Yonghui Wu, Jean-Baptiste Alayrac,
  Jiahui Yu, Radu Soricut, Johan Schalkwyk, Andrew~M Dai, Anja Hauth, et~al.
  2023.
\newblock Gemini: a family of highly capable multimodal models.
\newblock \emph{arXiv preprint arXiv:2312.11805}.

\bibitem[{Thirunavukarasu et~al.(2023)Thirunavukarasu, Ting, Elangovan,
  Gutierrez, Tan, and Ting}]{thirunavukarasu2023large}
Arun~James Thirunavukarasu, Darren Shu~Jeng Ting, Kabilan Elangovan, Laura
  Gutierrez, Ting~Fang Tan, and Daniel Shu~Wei Ting. 2023.
\newblock Large language models in medicine.
\newblock \emph{Nature medicine}, 29(8):1930--1940.

\bibitem[{Touvron et~al.(2023{\natexlab{a}})Touvron, Lavril, Izacard, Martinet,
  Lachaux, Lacroix, Rozi{\`e}re, Goyal, Hambro, Azhar
  et~al.}]{touvron2023llamaa}
Hugo Touvron, Thibaut Lavril, Gautier Izacard, Xavier Martinet, Marie-Anne
  Lachaux, Timoth{\'e}e Lacroix, Baptiste Rozi{\`e}re, Naman Goyal, Eric
  Hambro, Faisal Azhar, et~al. 2023{\natexlab{a}}.
\newblock Llama: Open and efficient foundation language models.
\newblock \emph{arXiv preprint arXiv:2302.13971}.

\bibitem[{Touvron et~al.(2023{\natexlab{b}})Touvron, Martin, Stone, Albert,
  Almahairi, Babaei, Bashlykov, Batra, Bhargava, Bhosale
  et~al.}]{touvron2023llamab}
Hugo Touvron, Louis Martin, Kevin Stone, Peter Albert, Amjad Almahairi, Yasmine
  Babaei, Nikolay Bashlykov, Soumya Batra, Prajjwal Bhargava, Shruti Bhosale,
  et~al. 2023{\natexlab{b}}.
\newblock Llama 2: Open foundation and fine-tuned chat models.
\newblock \emph{arXiv preprint arXiv:2307.09288}.

\bibitem[{Wei et~al.(2022)Wei, Wang, Schuurmans, Bosma, Xia, Chi, Le, Zhou
  et~al.}]{wei2022chain}
Jason Wei, Xuezhi Wang, Dale Schuurmans, Maarten Bosma, Fei Xia, Ed~Chi, Quoc~V
  Le, Denny Zhou, et~al. 2022.
\newblock Chain-of-thought prompting elicits reasoning in large language
  models.
\newblock \emph{Advances in neural information processing systems},
  35:24824--24837.

\bibitem[{Xiao et~al.(2022)Xiao, Chen, Guo, Zhuang, and
  Wang}]{xiao2022decoupled}
Teng Xiao, Zhengyu Chen, Zhimeng Guo, Zeyang Zhuang, and Suhang Wang. 2022.
\newblock Decoupled self-supervised learning for graphs.
\newblock \emph{Advances in Neural Information Processing Systems},
  35:620--634.

\bibitem[{Xiao et~al.(2021)Xiao, Chen, Wang, and Wang}]{xiao2021learning}
Teng Xiao, Zhengyu Chen, Donglin Wang, and Suhang Wang. 2021.
\newblock Learning how to propagate messages in graph neural networks.
\newblock In \emph{Proceedings of the 27th ACM SIGKDD Conference on Knowledge
  Discovery \& Data Mining}, pages 1894--1903.

\bibitem[{Yang et~al.(2022)Yang, Hu, Babuschkin, Sidor, Liu, Farhi, Ryder,
  Pachocki, Chen, and Gao}]{yang2022tensor}
Greg Yang, Edward~J Hu, Igor Babuschkin, Szymon Sidor, Xiaodong Liu, David
  Farhi, Nick Ryder, Jakub Pachocki, Weizhu Chen, and Jianfeng Gao. 2022.
\newblock Tensor programs v: Tuning large neural networks via zero-shot
  hyperparameter transfer.
\newblock \emph{arXiv preprint arXiv:2203.03466}.

\bibitem[{Yang et~al.(2023)Yang, Yu, Zhu, and Hayou}]{yang2023tensor}
Greg Yang, Dingli Yu, Chen Zhu, and Soufiane Hayou. 2023.
\newblock Tensor programs vi: Feature learning in infinite-depth neural
  networks.
\newblock \emph{arXiv preprint arXiv:2310.02244}.

\bibitem[{Yuksel et~al.(2012)Yuksel, Wilson, and Gader}]{yuksel2012twenty}
Seniha~Esen Yuksel, Joseph~N Wilson, and Paul~D Gader. 2012.
\newblock Twenty years of mixture of experts.
\newblock \emph{IEEE transactions on neural networks and learning systems},
  23(8):1177--1193.

\bibitem[{Yun et~al.(2024)Yun, Zhuang, Fu, Xing, and Zhang}]{Yun2024TowardIM}
Longfei Yun, Yonghao Zhuang, Yao Fu, Eric~P. Xing, and Hao Zhang. 2024.
\newblock \href {https://api.semanticscholar.org/CorpusID:268875826} {Toward
  inference-optimal mixture-of-expert large language models}.
\newblock \emph{ArXiv}, abs/2404.02852.

\end{thebibliography}
\appendix

\section{Experimental Settings}\label{sec:appendix1}
The architecture of all models (including Dense Models and MoE Models) are shown in table \ref{tab:dense_arch} and table \ref{tab:moe_arch}, respectively. The number of layers, attention heads, hidden dimensions, and other relevant details are listed in the tables.

During training, we employed the AdamW optimizer with parameters \(\beta_1 = 0.9\) and \(\beta_2 = 0.95\) for all models. Following the Chinchilla law \citep{hoffmann2022training}, we established a maximum learning rate of \(1.5 \times 10^{-3}\) for smaller models and \(2 \times 10^{-4}\) for larger ones. A cosine scheduler with a 10x learning rate decay was implemented throughout the training process. We applied Gaussian smoothing with a 10-step window length to enhance the training curve.

\begin{figure}[t]
	\centering
	\vspace{-0.1in}        \includegraphics[width=0.98\linewidth,height=2.3in]{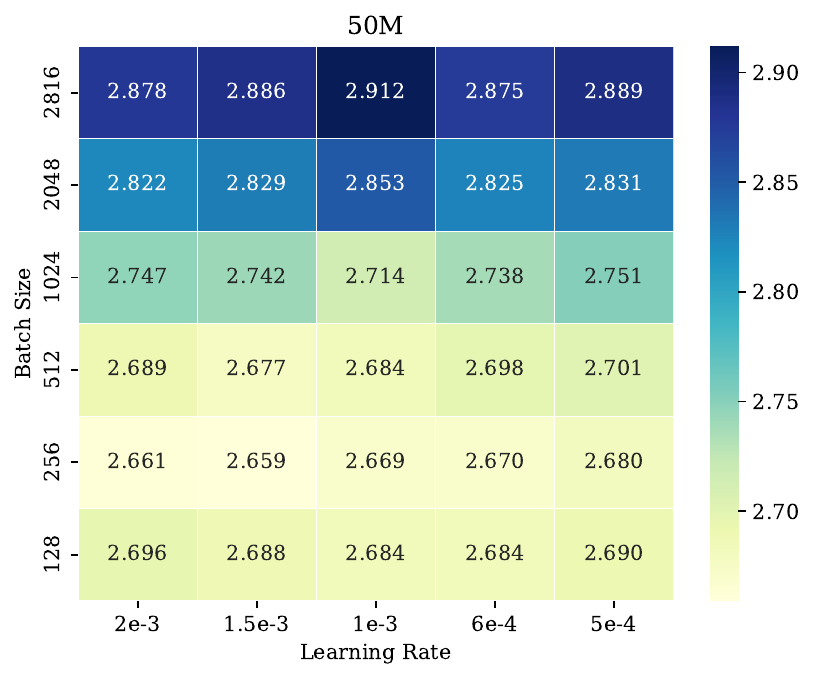}	
	\vspace{-0.1in}	
	\caption{We observed that the optimal settings are a Learning Rate of 1.5e-3 and a Batch Size of 256. Additionally, in the neighborhood of these settings, the training loss values have an error within 2\%, indicating that the model will converge to the optimal value around the best hyperparameter settings.}
	\vspace{-0.1in}
	\label{fig:batch_size_learning_rate_50m}
\end{figure}

Specifically, we identified the best performance values within our hyperparameter range. The range of hyperparameter settings, including batch size and learning rate, was carefully selected for each model size to ensure optimal performance within the designated FLOP budget. Our observations indicate that performance tends to converge to optimal values around the neighborhood of the best settings, as illustrated in Figure~\ref{fig:batch_size_learning_rate_50m}.

The dataset we used for the training of Dense Models and MoE Models is The Pile \citep{2021arXiv210100027G}, which is a 825 GiB English text corpus consisting of 22 high-quality subsets.

\begin{table}[htbp]
\centering 
\fontsize{6.5}{6.5}\selectfont
\caption{Details about the architecture of our Dense Models involved in all experiments.}
\label{tab1}
\renewcommand{\arraystretch}{2}
\begin{tabular}{cccccccc}
\toprule
Model Size & Params. & Hid. Size & Layers & Head Size & FFN\\
\midrule
50M & 47.71M & 512 & 14 & 4 & 1536\\
100M & 113.25M & 768 & 16 & 6 & 2048 & \\
500M & 487.59M & 1280 & 24 & 10 & 3584 \\
1B & 936.31M & 1664 & 28 & 13 & 4480 \\
7B & 7.03B & 4096 & 32 & 32 & 11008\\
\bottomrule
\label{tab:dense_arch}
\end{tabular}
\end{table}

\begin{table}[h]
\centering
\fontsize{6.5}{6.5}\selectfont
\caption{Details about the architecture of our MoE Models involved in all experiments, all of them are top-3.}
\label{tab1}
\renewcommand{\arraystretch}{2}
\begin{tabular}{ccccccccc}
\toprule
Model Size & Act. Params & Hid. Size & Layers & Head Size & Int. Size\\
\midrule
200M & 183M & 640 & 12 & 8 & 1792\\
700M & 692M & 1280 & 16 & 16 & 3456\\
1.5B & 1.45B & 1600 & 24 & 20 & 4352 \\
3B & 3.0B & 2304 & 28 & 32 & 6272 \\
\bottomrule
\end{tabular}
\label{tab:moe_arch}
\end{table}

\section{Discussion of related works}

Large language models (LLMs) have obtained significant attention and undergone substantial development in recent years. They can be categorized into two main classes based on their parameter utilization during the forward pass: Dense Models and MoE (Model of Experts) Models. Dense Models, such as GPT-2 \citep{Radford2019LanguageMA}, GPT-3 \citep{brown2020language}, Llama \citep{touvron2023llamaa, touvron2023llamab}, Chinchilla \citep{hoffmann2022training}, and Gopher \citep{rae2021scaling}, maintain a total parameter count equal to the active parameter count per forward pass. On the other hand, MoE Models \citep{shazeer2017outrageously} activate only a subset of total parameters during training, as seen in models such as Mixtral 8x7B \citep{jiang2024mixtral}, Switch Transformer \citep{fedus2022switch}, GShard \citep{lepikhin2020gshard}, GLaM \citep{du2022glam}, and DeepSpeed-MoE \citep{rajbhandari2022deepspeed}. Dense Models are known for their simplicity in implementation and training. However, MoE Models can scale to significantly larger total parameter counts without a proportional increase in computational cost. Despite challenges like load balancing and expert selection faced by MoE Models, prior research indicates that they offer superior performance due to increased model capacity and enhanced data efficiency.

Due to the significant costs associated with training process, understanding the scaling laws of large language models (LLMs) is crucial. Some previous studies \citep{kaplan2020scaling, bahri2021explaining} have proposed a power-law relationship between loss and various factors like the number of non-embedding parameters, training tokens, and compute budget across different magnitudes. Notably, \citet{kaplan2020scaling} found that increasing the model size by 8 times only requires a roughly 5x increase in data to avoid penalties. In contrast to earlier findings, \citet{hoffmann2022training} implements optimized training configurations, which include the use of training tokens and learning rate schedules, and recommends scaling training tokens in proportion to model size. Additionally, research by \citet{bi2024deepseek} explores the scaling laws of batch size and learning rate in relation to model scale (non-embedding FLOPs per token), offering a more precise estimation. They propose an allocation strategy for scaling up models and data based on dataset quality. While some studies \citep{li2024surge, mccandlish2018empirical} link the optimal batch size to gradient noise scale and optimizer type, recent attention has shifted towards Mixture of Expert (MoE) models due to their potential cost savings. \citet{fedus2022switch} investigates the scaling properties of MoE Models and suggest that having more parameters (experts) with a fixed computational budget accelerates training and surpasses dense Transformer baselines. On the other hand, \citet{Yun2024TowardIM} incorporate the hyperparameter $E$ (number of experts) into existing scaling laws, revealing diminishing returns with increasing expert numbers. However, a systematic investigation into the scaling laws of MoE Models' hyperparameters and the transferability of scaling laws between Dense Models and MoE Models remains lacking, which will be the focus of our work.

With the increasing size of models in deep learning, hyperparameter estimation has gained significant attention due to the huge costs associated with hyperparameter tuning. In training large language models, several key hyperparameters require careful consideration and selection. Previous research has offered valuable insights that inspire our approach. For instance, \citet{mccandlish2018empirical} highlights the trade-off between training speed and efficiency, focusing on the critical batch size, which can be measured by the gradient noise scale. Other studies \citep{goyal2017accurate, granziol2022learning, li2024surge} propose a linear scaling rule for adjusting learning rates as a function of minibatch size for SGD optimizers, while a square root scaling rule is suggested for adaptive optimizers. Some works focus on optimal hyperparameter transfer. For instance, unlike traditional Bayesian Optimization (BO)-based methods \citep{horváth2021hyperparameter, Perrone2018ScalableHT}, recent studies \citep{yang2022tensor, yang2023tensor} emphasize transferring hyperparameters across model scales. They introduce a novel zero-shot strategy known as Maximal Update Parametrization ($\mu$P), demonstrating that optimal hyperparameter choices for smaller models remain effective for larger ones. However, there is a lack of research on hyperparameter transfer across MoE Models and Dense Models. Our work specifically emphasizes critical hyperparameters transfer such as resource allocation, learning rate, and batch size, and their transfer rules for Dense Models and MoE Models.

\section{Details of Differentiation}

Given the loss function:
\begin{equation}
    \hat{L}(N, D, E) = \frac{A}{N^\alpha E^\gamma} + \frac{B}{D^\beta} + \sigma
    \label{eq:loss_N_D_E_appendix}
\end{equation}
and the constraint:
\begin{equation}
   C = N D
   \label{eq:constraint}
\end{equation}

\noindent where $N$, $D$, $C$ and $E$ are the model scale (non-embedding FLOPs per token) and the number of training tokens, compute budget and the number of experts respectively. $A$, $B$, $\alpha$, $\beta$, $\gamma$ are all coefficients. $\sigma$ is the noise of the dataset, which is the minimum loss value for model. 

We can substitute \(D = \frac{C}{N}\) into the loss function. Next, we differentiate \(\hat{L}\) with respect to \(N\):
\begin{equation}
   \hat{L}(N, \frac{C}{N}, E) = \frac{A}{N^\alpha E^\gamma} + \frac{B N^\beta}{C^\beta} + \sigma
\end{equation}

Next, we differentiate \(\hat{L}\) with respect to \(N\):
\begin{equation}
    \frac{d \hat{L}}{d N} = \frac{d}{d N} \left( \frac{A}{N^\alpha E^\gamma} + \frac{B N^\beta}{C^\beta} + \sigma \right)
\end{equation}

The $N_{\text{opt}}$ and $D_{\text{opt}}$ are the critical point when $L$ achieves the minimum loss value. Set the derivative to zero to find the critical points:
\begin{equation}
    -\frac{\alpha A}{N^{\alpha + 1} E^\gamma} + \frac{\beta B N^{\beta - 1}}{C^\beta} = 0
\end{equation}

Therefore,
\begin{equation}
    N_{\text{opt}}(C) = \left( \frac{\alpha A}{\beta B E^\gamma} \right)^{\frac{1}{\alpha + \beta}} C^{\frac{\beta}{\alpha + \beta}}
\end{equation}

Using the constraint \(C = N D\), we find \(D_{\text{opt}}\):
\begin{equation}
    D_{\text{opt}}(C) = \frac{C}{N_{\text{opt}}}
\end{equation}

\begin{equation}
   D_{\text{opt}}(C) = \frac{C}{\left( \frac{\alpha A}{\beta B E^\gamma} \right)^{\frac{1}{\alpha + \beta}} C^{\frac{\beta}{\alpha + \beta}}}
\end{equation}

Therefore,
\begin{equation}
    D_{\text{opt}}(C) = \left( \frac{\alpha A}{\beta B E^\gamma} \right)^{- \frac{1}{\alpha + \beta}} C^{\frac{\alpha}{\alpha + \beta}}
\end{equation}

\end{document}